\documentclass[conference]{IEEEtran}
\usepackage{graphicx}
\usepackage{amsmath}
\usepackage{mathtools}
\usepackage{latexsym}

\usepackage{amsthm}
\theoremstyle{definition}
\newtheorem{theorem}{Definition}
\newtheorem*{theorem*}{Definition}
\newtheorem{definition}[theorem]{Definition}
\newtheorem*{definition*}{Definition}

\usepackage{listings}

\usepackage{algorithm}
\usepackage{algorithmic}

\begin{document}

\title{A Curious New Result of Resolution Strategies in Negation-Limited Inverters Problem}

\author{

\IEEEauthorblockN{Ruo Ando}
\IEEEauthorblockA{
National Institute of Informatics\\
2-1-2 Hitotsubashi, Chiyoda-ku, Tokyo \\101-8430 Japan\\
}
\and
\IEEEauthorblockN{Yoshiyasu Takefuji}
\IEEEauthorblockA{Keio University\\
5322 Endo Fujisawa, Kanagawa \\252-0882 Japan\\
}
}

\maketitle

\begin{abstract}
Generally, negation-limited inverters problem is known as a puzzle of constructing an inverter with AND gates and OR gates and a few inverters.
In this paper, we introduce a curious new result about the effectiveness of two powerful ATP (Automated Theorem Proving) strategies on tackling negation limited inverter problem. 
Two resolution strategies are UR (Unit Resulting) resolution and hyper-resolution.
In experiment, we come two kinds of automated circuit construction: 3 input/output inverters and 4 input/output BCD Counter Circuit.
Both circuits are constructed with a few limited inverters. 
Curiously, it has been turned out that UR resolution is drastically faster than hyper-resolution in the measurement of the size of SOS (Set of Support). 
Besides, we discuss the syntactic and semantic criteria which might causes considerable difference of computation cost between UR resolution and hyper-resolution.
\end{abstract}

\IEEEpeerreviewmaketitle

\section{Introduction}
A circuit with outputs $ \lnot x_1, \lnot x_2, ... , \lnot x_n $ for any Boolean inputs $ x_1, x_2, ... x_n $ is called as an inverter. 
Here, we consider the automated construction of some circuits with AND gates and OR gates and a few NOT gates. 
This is also called as a knotty problem in \cite{Sallows}.
Before \cite{Sallows}, Larry Wos \cite{Wos1} introduced ``two-inverter puzzle'' in an article on Automated reasoning. 
One can easily construct an inverter by placing n NOT gates in a row. 
In two-inverter puzzle, we have constrain that an inverter using fewer than n NOT gates and with to construct an inverter with an arbitrary number of AND gates and OR gates. 
Theoretically, it is known that [log(n+1)] NOT gates are necessary and sufficient to construct the inverter \cite{Sheldon}. 

\begin{figure}[ht]
\centering
\includegraphics[scale=0.3]{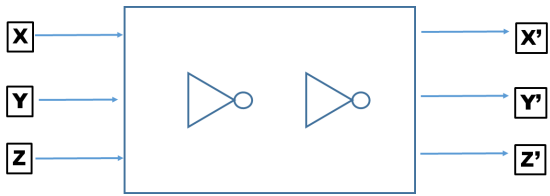}
\caption{Two inverters puzzle as negation-limited inverters problem.}
\end{figure}

Figure 1 the depicts 3-input/3-output version of this problem which is called as two-inverter puzzle.
The black box in the middle of the figure receives binary inputs [0,1] at the terminal of x, y and z.
Each output terminals of x', y' and z' yields the complement of the corresponding input. 
For example, if x is 0, x' is 1 and so on. 
For three terminals, each of the eight possible 3-bit input words and therefore yields its complementary word at the outputs.
Normally, without any constraint about negation-limited inverters, 
such a transfer function would be implemented by using three inverters 
or NOT-gates connected between each input and output.

In formal, this puzzle is designed as design a network using any number of AND and OR gates, but not more than two (2) NOT's 
to achieve exactly the same input-output function. The AND's and OR's may have as many inputs as required.

To discuss a curious new result of this paper mainly concerning two-inverter puzzle, we define the broader problem as follows.

\begin{definition}
\underline{Definition of negation limited inverters problem.} As a generalization of two-inverter puzzle, 
we define negation limited inverters problem as the construction of arbitrary N-bit input / N-bit output circuit with limited inverters of N-1.
\end{definition}

In \cite{Sheldon}, it was proved that the complete set of input variables may be inverted D(n) inverters where D(n) is the small integer y such that $n < 2^y$.
According to this proof, the negation limited inverters problem, including two inverter puzzle, includes also the constructing BCD (decade counter circuit) with 2 inverters \cite{Brcic} discussed in the following section.

\section{OTTER}

\subsection{OTTER and its clause sets}
The theorem prover OTTER (Organized Techniques for Theorem-proving and Effective Research) has been developed by W. McCune as a product of Argonne National Laboratory.
OTTER is based on earlier work by E. Lusk, R. Overbeek and others \cite{Lusk2}.
By the research efforts of \cite{McCune1} \cite{McCune2} \cite{Lusk} for certain classes of problem, 
OTTER is widely regarded as the most powerful automated deduction system. 
OTTER adopts the given-clause algorithm and implements the set of support strategy \cite{Wos2}.

In given-clause algorithm, all retained clauses are divided into two sets.
The first sets are called as the set of support (SOS). OTTER starts with the retention of a set of support including all of the choses input
clauses. During the run, the initial set of support and the clauses which are generated are retained.
The second set is the usable list. At beginning phase of reasoning, the usable list is not included into the initial set of support. 
The usable list are the clauses which were once in the set of support but have been already picked up as the focus of attention for deducing additional clauses.
More technically, in detail, OTTER maintains four lists of clauses in reasoning process.

\begin{enumerate}
\item Usable. This list works as a rule by keeping clauses which are available to make inferences.

\item SoS. Clauses is regarded as facts. Set of support are not used to make inferences. They are kept to participate in the search.

\item Passive. They are specified to be used only for forward subsumption and unit conflict. The passive list does not participate in the search.
The passive list does not change from the start of reasoning process as fixed input.

\item Demodulators. Demodulators are used to rewrite newly inferred clauses with equalities.
\end{enumerate}

In this paper, particularly, we focus on the size of set of support list. 
Set of support is important indicator for introspecting the reasoning process. 

\subsection{Given Clause algorithm}

OTTER adopts given-clause algorithm in which the program attempts to use any and all combinations from axioms in given clause. 
In other words, the combinations of clause are generated from given clauses which has been focused on.

\begin{algorithm}
\caption{Given clause algorithm}
\label{alg1}
\begin{algorithmic}[1]
\renewcommand{\algorithmicrequire}{\textbf{Input:}}
\renewcommand{\algorithmicensure}{\textbf{Output:}}
\REQUIRE SOS, Usable List
\ENSURE Proof
\WHILE{until SOS is empty}
\STATE choose a given clause G from SOS;
\STATE move the clause g to Usable List;
\WHILE { c\_1, ..., c\_n in Usable List} 
\WHILE{$ R(c_1,..c_i, G ,c_{i+1},..c_n) exists $}
\STATE $ A \Leftarrow R(c_1,..c_i, G, c_{i+1},..c_n); $
\IF{A is the goal}
\STATE report the proof;
\STATE stop
\ELSE[A is new odd]
\STATE add A to SOS X
\ENDIF
\ENDWHILE
\ENDWHILE
\ENDWHILE
\end{algorithmic}
\end{algorithm}

At line 2, given clause G is extracted from SOS (Set of Support). 
Line 4 and 5 is a loop to use any and all combinations of given clause and Usable List.
In detail, \cite{Slaney} \cite{Graf} discuss the basic framework of given clause algorithm.
To put it simply, given clause algorithm consists of the following steps.

\begin{enumerate}
\item Pick up a clause (called the given clause) from the set of support.
\item Add the given clause to the usable list.
\item Applying the inference rule or rules in the effect, infer all clauses which are generated from the given clause (one parent) and the usable list (other parents). 
\item Process newly inferred clause. 
\item Append each inferred new clause to the SoS. These clause is not discarded as a result of processing. Exactly, this is done in the course of processing the newly generated clause.
\end{enumerate}

In a nutshell, the reasoning program chooses a clause from the clauses which is focused on in the set of support.
The selected clause is called as focal clause or given clause.

\section{Resolution strategy}

\subsection{Hyper-resolution}

The hyper-resolution inference rule is most frequently used one. Hyper-resolution takes a non-positive clause which is called as the nucleus and simultaneously 
infers each of its negative literals. Those negative literals are called as the satellites. 
Hyper-resolution can be regarded as a sequence of positive binary resolution steps yielding a positive clause.

\begin{definition}
\underline{Definition of Hyper-resolution.} 
The inference rule hyper-resolution, originally introduced by \cite{Robinson}, processes simultaneously a clause which contains
at least one negative literal and a set of clauses $A[i]$, each of which contains only positive literals. 
Then, hyper-resolution yields a clause B containing only one positive literals when successful.
The clause B is obtained by finding an MGU (most general unifier) which is also denoted as $ \sigma $. MGU ($ \sigma $)
simultaneously unifies one positive literal in each of the $A[i]$ with a negative literal $NL$ and is applied to yields $NL'$ and $A[i]'$.
The clauses $NL$ and $A[i]$ are assumed pairwise to have no variables in common.
The $NL$ is called as the nucleus. And the clause $A[i]$ are term the satellites.
\end{definition}

The general scheme is:

\begin{align*}
& {K_{1,l},.,K{1,n}} \\
& ...\\
& {K{m,l},.,K{m,n}}\\
& \underline{ \{\lnot L_1, . , \lnot L_{m+1}, L_{l}\} \exists \sigma.\sigma} \\
& \underline{ = mgu( [ | K_1 |, . , | K_{m,1} |], [|L_1|, . , |L_m|]) } \\
& \{K_{1,2},.,K_{1,n},K_{m,2},.K_{m,n},L_{m+1},.,L_l \} \sigma 
\end{align*}

In the list above, $K_i$ denotes clause and $L_i$ denotes literal.
Hyper-resolution is applied to a set of m unit clauses {K1} ... {Km} and a single nucleus {L1, ..., Lm+1} consisting of m + 1 literals.

Roughly, corresponding to OTTER's syntax, the ``if-then'' clause may have more than one conclusion literal.

\begin{verbatim}
1: If P & Q then R | S
\end{verbatim}

deduces

\begin{verbatim}
1: -P | -Q |-R | S
\end{verbatim}

Then, the process of clash are occurred under the he hypothesis literals in the ``if-then'' with more than one literal.
A typical pattern might be as follows:

\begin{verbatim}
1: -P | -Q |-R | S
2: P | T
3: Q | W
4: R
5: -------
6: T | W | S.
\end{verbatim}

Importantly, hyper-resolution requires that all of the negative literals in the ``if-then'' clause be clashed with corresponding literals in other clauses.
For example, from 

\begin{verbatim}
1: -P(x,y) | -Q(x) |-R(x,y) 
2: P(z,b)
3: Q(a)
\end{verbatim}

Hyper-resolution deduces

\begin{verbatim}
1: R(a,b).
\end{verbatim}

Hyper-resolution is most frequently adopted inference rule in the situations where equality substitutions do not play a major role.
Hyper-resolution is intuitively natural to human reasoning. 

According to \cite{Lusk2}, ``Don't draw any conclusions until all of the hypothesis are satisfied'' is the restriction 
which all negative literals should be clashed.

In general, for a broad class of reasoning problem, hyper-resolution is sufficient. 
It is the rule that most resembles the inference mechanism used in deduction systems. 
Also, In OTTER, hyper-resolution is the default inference rule.

\subsection{UR-Resolution}
The UR-resolution (unit-resulting resolution) inference rule \cite{McCharen} takes a non-unit clause to produce a unit clause.
In UR resolution, the non-unit clause is called as the nucleus. The unit clause is called as the satellites.
The satellites are clauses to be resolved all but one of its literals with unit clauses.
Further, UR resolution is divided into two. Positive UR-resolution has the constraint that the result must be a positive unit clause
while the constraint of negative UR-resolution is that the result must be a negative unit clause.

The general scheme is:
\begin{align*}
& {K_1} \\
& ...\\
& {K_m}\\
& \underline{ \{L_1, . , L_{m+1}\} \exists \sigma.\sigma = mgu( [ | K_1 |, . , | K_m |], [|L_1|, . , |L_m|]) } \\
& \{L_{m+1} \} \sigma 
\end{align*}

In the list above, $K_i$ denotes clause and $L_i$ denotes literal.

UR-resolution inference rules take a set of $ m $ unit clauses ${K1} ... {Km}$ and a single nucleus ${L1, ..., Lm+1}$ consisting of $m + 1$ literals.
Here, $ K_i, L_i $, then $ L_{m+1} \sigma $ is called as the unit resulting resolvent.
In the general scheme, all pairs of literals $ K_i, L_i $ should be complementary.
That is, $ K_i, L_i $ are assumed to have opposite signs. 
Because $|K1|$ denotes the atom contained in the literal $K1$, reasoning process of the simultaneous unifier avoids the signs of the literals.

\begin{definition}
\underline{Definition of UR-resolution.} 
UR-resolution takes each literal to be removed from the nucleus. Then, taken literals are unified with a unit satellite. 
In UR-resolution, both negative and positive resolvents are supported. UR-resolution is not refutation complete. 
However, UR-resolution is refutation complete in coping with horn clause sets.
\end{definition}

Hyper-resolution will reach out to the derivation with only positive literalism. It is sufficient in coping with a large clauses of problem.
Instead of avoiding all restrictions on all clauses to be derived, UR-resolution consider the possibility of clauses containing a single literal.
Such clauses are called as unit clauses or simply units. 
In UR-resolution, a unit clause can be described as a statement of fact.
On the other hand, multi-literal clauses represent conditional statements in the case that multi-literal clauses contain both positive and negative literals.
Consequently, unit clauses are effective in many situations. UR-resolution discard the restriction that derived clauses should
have only positive literals. At the same time, UR-resolution imposes the restriction which derived clauses should be units. For example, 
.
\begin{verbatim}
1: -P | -Q | R
2: P
3: -R
\end{verbatim}

UR-resolution derives -Q. Note that hyper-resolution would be unable to derive anything on the contrast.
Besides, UR-resolution focuses on unit in a way which all but one of the clauses to participate in the deduction. 
Those clauses should be unit clauses although they can be either positive or negative. 
Broadly, UR-resolution focuses on unit clauses whereas hyper-resolution emphasizes positive clauses.

\subsection{Set of Support}

The set of support strategy \cite{Wos2} guides the reasoning program to select from the clauses characterizing the question under research to be put in the initial set of support list which is denoted as $ list(sos)$ in OTTER.
The corresponding restriction prevents the reasoning program from adopting an inference rule to a set of clauses of which all clauses are complement of the set of support.
Consequently, each clause generated and retained is appended to $list(sos)$.
In \cite{Wos2}, experimentally, it is pointed out that the most effective choice for the initial set of support is based on the special hypothesis and the denial of the theorem under research.
The second best choice is the denial of the theorem itself.

\begin{definition}
\underline{Definition of Set of Support Strategy.} Let $R$ be any inference rules, and let $S$ be any nonempty set of clauses.
The set of support strategy requires selecting a nonempty subset $T of S$. Let $T_0$ be the set of all clauses D. 
Here, D is in T or D is a factor of a clause C in T. 
Also, let $T_1$ be the set of clauses D.
Here, D is deduced by applying $R$ to the set $C_1, C_2, ... C_n$ with at least one of $C_j$ in $T_0$.
$C_k$ are not in $T_0$. Or a factors of clause in $S$ such that $D$ is a factor of a clause in $T_1$.
\end{definition}

In negation limited inverters problem (two-inverter puzzle), the set of support contains the statements which the input signals are constructible.

\begin{center}
\begin{lstlisting}[caption = Set of Support list, label = program1]
P(00001111, v). input 1
P(00110011, v). input 2
P(01010101, v). input 3
\end{lstlisting}
\end{center}

We then add a statement denying that the puzzle can be solved. 
The denial says that at least one of the desired output patterns cannot be constructed.

\begin{center}
\begin{lstlisting}[caption = Usable list, label = program1]
-P(0000000110, v) | 
-P(0001111000, v) | 
-P(0110011000, v) | 
-P(1010101010, v).
\end{lstlisting}
\end{center}

Reasoning program is terminated when the unit conflict occur. 
Unit conflict is an inference rule that derives a contradiction from unit clauses.
For example, unit conflict occurs between P(a,b) and -P(x,b).
Theoretically, unit conflict is based on proof by contradiction.


\section{Experiment}

In this section we describe the experimental results of the training and generating sine wave. 
In experiment, we use workstation with Intel(R) Xeon(R) CPU E5-2620 v4 (2.10GHz) and 251G RAM.

\subsection{Tracking the size of set of support}

The main loop for inferring and processing clauses and searching for a refutation operates
mainly on the lists usable and sos.

\begin{enumerate}
\item Choose appropriate $given\_clause$ in sos;
\item Move $given\_clause$ from $list(sos)$ to $list(usable)$
\item Infer and process new clauses using the inference rules set. 
\item Newly generated clause must have the $given\_clause$. 
\item Do the retention test on new clauses and append those to $list(sos)$.
\end{enumerate}

Main loop is depicted in Algorithm 3.

\begin{algorithm}
\caption{Tracking the size of set of support}
\label{alg1}
\begin{algorithmic}[1]
\WHILE{given clause is NOT NULL}
\STATE $ index\_lits\_clash(giv\_cl); $
\STATE $ append\_cl(Usable, giv\_cl); $
\IF{$ splitting() $ }
\STATE $ possible\_given\_split(giv\_cl); $
\ENDIF
\STATE{infer\_and\_process(giv\_cl);}
\STATE{giv\_cl = extract\_given\_clause();}
\STATE{track(sos\_size);}
\ENDWHILE
\end{algorithmic}
\end{algorithm}

At line 9, we track the size of set of support for each iteration step.
After line 8 of picking up the clause from set of support, we can record the current size of set of support. 
By doing this, we can obtain the plot with the iteration step of X-axis and the size of SOS of Y-axis as shown in the next section.

\subsection{Two inverter puzzle}

As an example of a logic circuit design problem, we consider the two-inverter puzzle.
In detail, using any number of AND, OR gates 
but no more than two NOT gates (inverters), construct a logic circuit with three inputs i1, i2, i3 
and three outputs o1, o2, o3.
Here, o1=NOT (il), o2=NOT (i2), o3=NOT (i3).
Table I shows input/output diagram of two-inverter puzzle.

\begin{table}[htb]
\begin{center}
\caption{Two inverter puzzle}
\begin{tabular}{|l|c|r|r|r|r|} \hline
\multicolumn{3}{|l|}{3-INPUTS} & \multicolumn{3}{|l|}{3-OUTPUTS}\\ \hline
0 & 0 & 0 & 1 & 1 & 1\\ \hline
0 & 0 & 1 & 1 & 1 & 0\\ \hline
0 & 1 & 0 & 1 & 0 & 1\\ \hline
0 & 1 & 1 & 1 & 0 & 0\\ \hline
1 & 0 & 0 & 0 & 1 & 1 \\ \hline
1 & 0 & 1 & 0 & 1& 0 \\ \hline
1 & 1 & 0 & 0 & 0 & 1 \\ \hline
1 & 1 & 1 & 0 & 0 & 0 \\ \hline
\end{tabular}
\end{center}
\end{table}


The first is that the circuit has three inputs. 

$ P(x1, x2, x3, x4, x5, x6, x7, x8). $ - (1)

This clause means it is possible to construct a circuit with the output signal pattern.
A complexity of this puzzle is how to keep track of how many inverters are used (only two are allowed in this case).
We can use a list for the notation. In the list, a variable is used to enable short list to subsume longer lists.
For example, the pattern (0,0,0,0,1,1,1,1) can be generated with no inverters, since it is one of the input signals. 

$ P(0,0,0,0,1,1,1,1,v). $ - (2)

If a signal pattern can be constructed by using a small number of inverters, then it does not matter if the same pattern can be generated by using one or more inverters.
If the signal is inverted, one inverter is added to the list, which can be described as:

$ P(1,1,1,1,0,0,0,0, L(inv(1,1,1,1,0,0,0,0), v)). $ - (3)

The inverter is denoted by the $inv()$ term. The term of $inv()$ captures the signal pattern at the output of inverter. 
If another inverter were used to the resulting signal pattern by reasoning program, this can be represented as:

\begin{verbatim}
P(0,0,0,0,1,1,1,1,
L(inv(0,0,0,0,1,1,1,1), 
L(inv(1,1,1,1,0,0,0,0),v)). - (4) 
\end{verbatim}

This clause (4) would be subsumed by the preceding two clauses (2) (3) immediately. 
And then two clauses (2)(3) are subsumed by (1) because the first clause (1) has the same pattern 
but has the empty inverter list.

Figure 2 shows the size of set of support during reasoning process of hyper-resolution.
X-axis is the iteration step. Y-axis is the size of set of support.
The size of SoS increases around the iteration step of 800. 
Then, it increases slowly in the next about 4,000 steps. 
After plateau from the iteration step 1000 to 4700, the SoS size increase speeds up until the iteration step 8000. 

Figure 3 depicts the size of set of support during reasoning process of UR-resolution.
The set size begins to increase rapidly about the iteration step of 600.
The speed of increase slows down around the iteration step around 1100.

\begin{figure}
\centering
\includegraphics[scale=0.5]{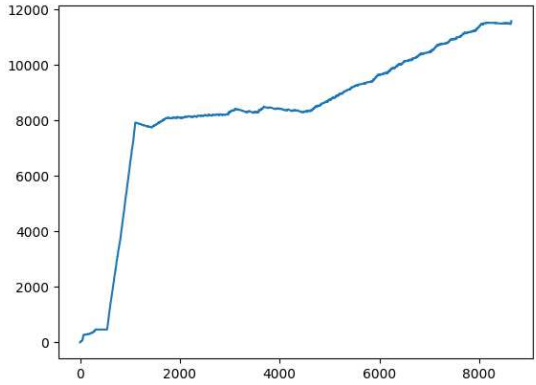}
\caption{Hyper-resolution in constructing 3-input/3-output inverter. X-axis of the number of iteration step. Y-axis is the size of set of support.}
\end{figure}

\begin{figure}
\centering
\includegraphics[scale=0.5]{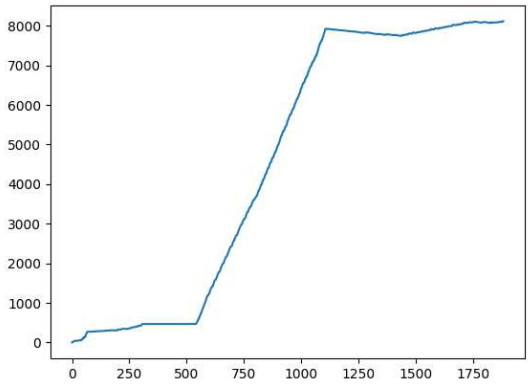}
\caption{UR-resolution in constructing 3-input/3-output inverter. X-axis of the number of iteration step. Y-axis is the size of set of support.}
\end{figure}

\subsection{BCD or Decade Counter Circuit}

A binary coded decimal (BCD) is a serial digital circuit designed for counting ten digits. BCD resets for every new input from clock. 
BCD is also called as ``Decade counter'' because BCD can go through 10 unique combinations of output.
With four digits, a BCD counter counts 0000, 0001, 0010, 1000, 1001, 1010, 1011, 1110, 1111, 0000, and 0001 and so on.

\begin{table}[htb]
\begin{center}
\caption{BCD (or Decade Counter Circuit)}
\begin{tabular}{|l|c|r|r|r|r|r|r|} \hline
\multicolumn{4}{|l|}{current state} & \multicolumn{4}{|l|}{next state}\\ \hline
0 & 0 & 0 & 0 & 0 & 0 & 0 & 1\\ \hline 
0 & 0 & 0 & 1 & 0 & 0 & 1 & 0\\ \hline
0 & 0 & 1 & 0 & 0 & 0 & 1 & 1\\ \hline
0 & 0 & 1 & 1 & 0 & 1 & 0 & 0 \\ \hline
0 & 1 & 0 & 0 & 0 & 1 & 0 & 1\\ \hline
0 & 1 & 0 & 1 & 0 & 1 & 1 & 0 \\ \hline
0 & 1 & 1 & 0 & 0 & 1 & 1 & 1\\ \hline
0 & 1 & 1 & 1 & 1 & 0 & 0 & 0\\ \hline
1 & 0 & 0 & 0 & 1 & 0 & 0 & 1 \\ \hline
1 & 0 & 0 & 1 & 1 & 0 & 1 & 0\\ \hline 
\end{tabular}
\end{center}
\end{table}

The table I describes the counting operation of Decade counter. It represents the count of circuit for decimal count of input pulses. The NAND gate output is zero when the count reaches 10 (1010).

\begin{figure}
\centering
\includegraphics[scale=0.5]{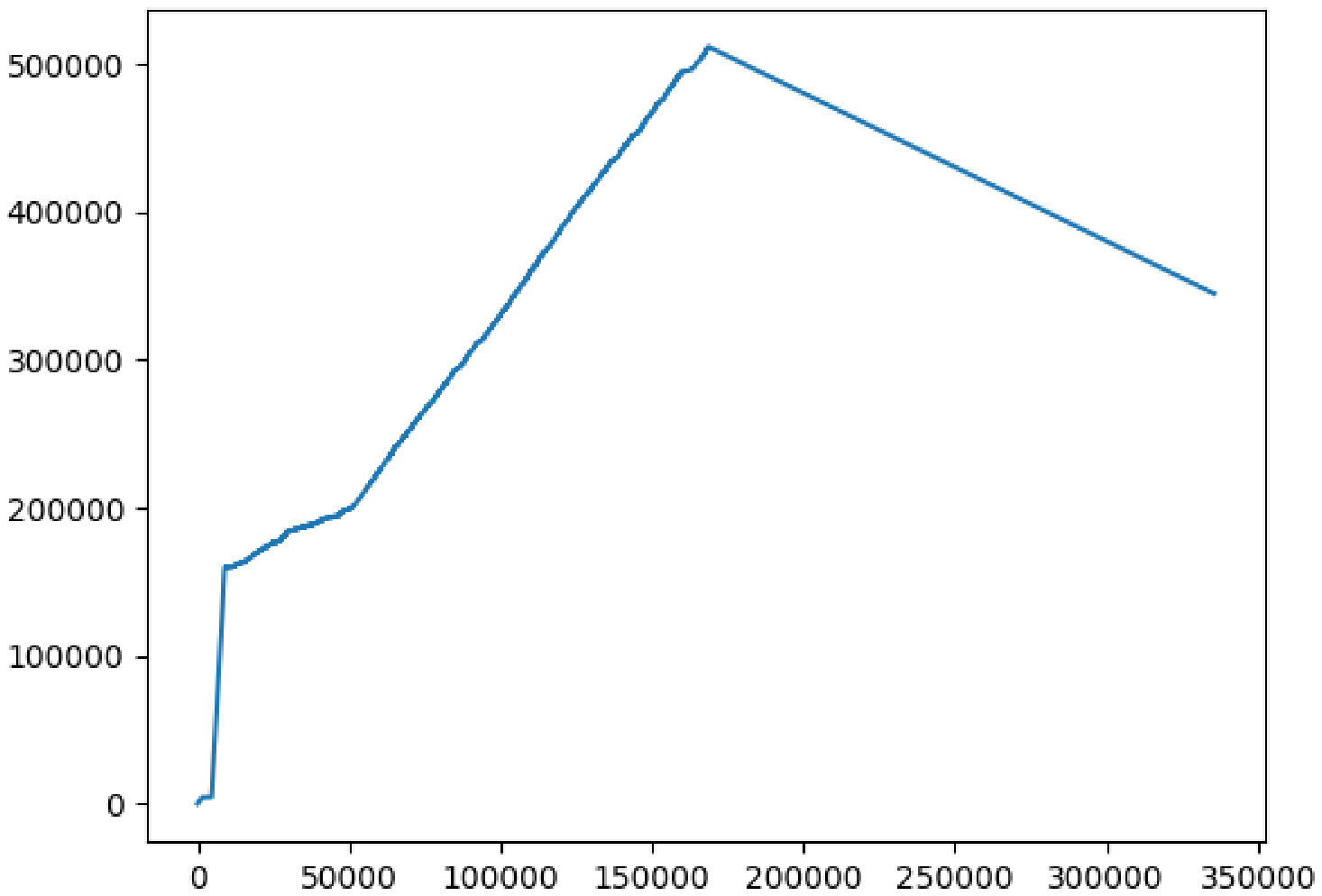}
\caption{Hyper-resolution for constructing BCD. X-axis of the number of iteration step. Y-axis is the size of set of support.}
\end{figure}

\begin{figure}
\centering
\includegraphics[scale=0.5]{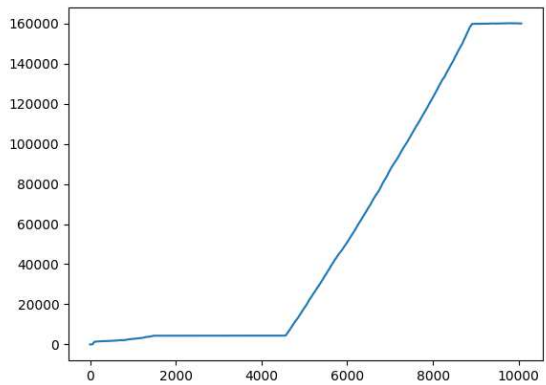}
\caption{UR-resolution for constructing BCD. X-axis of the number of iteration step. Y-axis is the size of set of support.}
\end{figure}

Figure 2 shows the size of set of support during reasoning process of hyper-resolution.
In detail, X-axis is the iteration step. 
The size of SoS increases rapidly from around the iteration step of 800. 
Then, it increases slowly in the next about 4,000 steps. 
After plateaus from the iteration step 1000 to 4700, the SoS size increase speeds up until the iteration step 8000. 

\subsection{Comparison}

In this section we illustrate the comparison of computation cost between hyper-resolution and UR-resolution.
In table III, concerning the number of clauses generated, UR resolution 53.68x faster than hyper-resolution.
Besides, UR-resolution takes 229.24 sec of user CPU time which is x113.35 faster than hyper-resolution.
On the contrast, the ratio of sos size of UR and hyper resolution is relatively small which is 21.55x.

\begin{table}[htb]
\begin{center}
\caption{BCD}
\begin{tabular}{|l|c|r|} \hline
& UR resplution & Hyper resolution \\ \hline
clauses generated & 5347949 & 287084274 \\ \hline
clauses forward subsumed & 5177890 & 286403743 \\ \hline
subsumed by sos & 1842716 & 44200930 \\ \hline
sos size & 160002 & 344900 \\ \hline
user CPU time & 229.45 (3min 49sec) & 25959.87 (7hr 12min) \\ \hline
\end{tabular}
\end{center}
\end{table}

Table IV shows the comparison between UR resolution and hyper-resolution in the construction of 3-input/3-output inverters.
The gap of the effectiveness between UR and hyper resolution is more moderate than the case of BCD.
Concerning the number of clauses generated, UR resolution 6.03x faster than hyper-resolution.
UR resolution takes less CPU user time than hyper-resolution by 6.34 times. 

\begin{table}[htb]
\begin{center}
\caption{Two inverter puzzle}
\begin{tabular}{|l|c|r|} \hline
& UR resplution & Hyper resolution \\ \hline
clauses generated & 342935 & 2069334 \\ \hline
clauses forward subsumed & 332937 & 2049117 \\ \hline
subsumed by sos & 77452 & 314773 \\ \hline
sos size & 8118 & 11577 \\ \hline
user CPU time (sec) & 3.99 & 25.31 \\ \hline
\end{tabular}
\end{center}
\end{table}

\section{Related work}
Switching theory was first introduced by Shannon \cite{Shannon} with the notable success in the practical application of Boolean algebra.
Knotty problem known as two inverter puzzle was first introduced by L.Wos \cite{Wos1}.
Sallows \cite{Sallows} discussed the negation-limited inverters problem in the viewpoint of Moore's original problem in circuit design and a seemingly analogous problem in computer programming.
Morizumi \cite{Morizumi} proposed the first negation-limited inverter of linear size using only o(n) NOT gates. 
In \cite{Sheldon}, it was shown that the complete set of input variables may be inverted D(n) inverters where D(n) is the small integer y such that $n < 2^y$.
In \cite{Tanaka}, Tanaka et al. proposed the construction of an inverter of size O(n log n) and depth O(log n) using [log (n+1)].

Originally, hyper-resolution was first illustrated in detail by \cite{Overbeek}.
L.Wos et al. discussed the efficiency and completeness of the set of support strategy in theorem proving in \cite{Wos2}.
Slaney et al. \cite{Slaney} proposed a model guided theorem prover which is called as SCOTT (Semantically Constrained Otter) with a resolution based automatic theorem proving.

Another powerful ATP (Automated Theorem Proving) strategies is base on equational reasoning \cite{Wos5}. 
Demodulation is regarded by many to be the inference rule to remove less obviously redundant information. 
It is designed to enable reasoning programs to simplify and canonicalize yielded clauses by applying demodulators which regarded as rewriting rules \cite{Wos6}.
Ando and Takefuji \cite{Ando2} proposes the application of demodulation for formal methods to analyze viral software metamorphism.
Wos proposes a look-ahead strategy which is called as hot list strategy to cope with the frequently occurred delay in focusing on a retained conclusion \cite{Wos7}.
A paramodulation inference rules is a basis of equational reasoning of OTTER. It consists of two parents and a child. 
By $from term$, the parent contains the equality for replacing literals. 
The $replaced term$ is called as into term. 
Paramodulation is regarded as the generalization of a substitution rule for equational reasoning. Paramodulation serves to build properties of equality along with demodulation.
Takefuji proposes the application of paramodulation to translator of Common Lisp \cite{Takefuji}.

Ando and Takefuji applies hot list strategy based on paramodulation for faster graph coloring \cite{Ando}. 
In \cite{Ando3}, hot list strategy is adopted for faster parameter detection of polymorphic viral code \cite{Ando}.

\section{Discussion}

\subsection{Hyper-resolution and UR-resolution}

Hyper-resolution inference rule has the advantage of coping with larger deduction steps than binary resolution does. 
On the other hand, hyper-resolution has the disadvantage of emphasizing syntactic criteria rather than semantic. 
UR-resolution inference rule has the advantage emphasizing semantic criteria, but disadvantages in taking certain types of 
problem. Particularly, UR-resolution requires all inferred clauses (conclusions) drawn into it to be unit clauses.
The clauses contain exactly one literal because unit clauses correspond to assertions rather than to a choice of possibilities.
Therefore, UR-resolution is semantically printed. 
Consequently, it is not hard to see that hyper-resolution inference rule leads to the derivation of clauses with only positive literals in them
whereas this is sufficient for a large clauses of problem, a number of reasoning tasks require the derivation of clauses containing negative literals.

\subsection{Set of support strategy}
Set of support strategy is the basis of all inference rules which OTTER adopts. 
As we discussed before, the set of support is the strategy to restrict the application of inference rules. 
Restriction strategies such as set of support and weighting are essential for fulfilling some given assignment in an feasible amount of computer time. 
If the reasoning program is not restricted properly, it will in almost all cases yields too (unacceptable) many conclusions. 

Currently, the set of support strategy is regarded as the most powerful restriction strategy available. 
Broadly, its use makes it possible to guide reasoning programs to prove theorems in far less computer time and memory than would be required.

Currently, the set of support strategy is considered by many to be the most powerful restriction strategy available.
In general, its use enables automated reasoning programs to prove theorems in far less computer time and memory than would be required as usual.
In \cite{Wos4}, a severe test of set of support strategy is provided in proving theorems relying the use of Godels finite axiomation of set theory.

\section{Conclusion}
In this paper, we discuss the novel result of resolution strategies in negation-limited inverter problem.
Two resolution strategies are UR (Unit Resulting) resolution and hyper-resolution, which makes the 
significant difference in computing time.

Particularly, in two-inverter puzzle, in the view of generated clauses, UR resolution 53.68x faster than hyper-resolution.
Besides, UR-resolution takes 229.24 sec of user CPU time which is x113.35 faster than hyper-resolution.
Curiously, it has been turned out that UR resolution is drastically faster than hyper-resolution in the measurements of the size of SOS (Set of Support).

For considering this novel result, we discuss the syntactic and semantic criteria which might causes considerable difference of computation cost between UR resolution and hyper-resolution.
Hyper-resolution will reach out to the derivation with only positive literalism. It is sufficient in coping with a large clauses of problem.
Instead of avoiding all restrictions on all clauses to be derived, UR-resolution consider the possibility of clauses containing a single literal.
For further work, we are going to inspect the detailed implementation of hyper-resolution and UR-resolution in detail.

\end{document}